\documentclass[conference]{IEEEtran}
\IEEEoverridecommandlockouts
\usepackage{cite}
\usepackage{amsmath,amssymb,amsfonts}
\usepackage{algorithmic}
\usepackage{algorithm}
\usepackage{graphicx}
\usepackage{textcomp}
\usepackage{comment}
\usepackage{float} 

\usepackage{tikz}
\usetikzlibrary{arrows.meta,positioning,shapes.geometric,calc,fit, shadows, backgrounds}

\tikzstyle{block} = [rounded rectangle, minimum width=1cm, minimum height=1cm, text centered, draw=black, thin, inner sep=2pt]
\tikzstyle{arrow} = [thick,->,>=stealth]

\usepackage{color}

\usepackage[utf8]{inputenc} 
\usepackage[T1]{fontenc}    
\usepackage{hyperref}       
\hypersetup{hidelinks}
\usepackage{url}            
\usepackage{booktabs}       
\usepackage{amsfonts}       
\usepackage{nicefrac}       
\usepackage{microtype}      
\usepackage{xcolor}         
\usepackage{enumitem}
\usepackage{comment}
\usepackage{tcolorbox}
\usepackage{listings}
\usepackage{pdfpages}
\usepackage{longtable}
\usepackage{multirow}
\usepackage{graphicx}
\usepackage{verbatim}
\usepackage{stfloats}
\usepackage[export]{adjustbox}

\usepackage{tabularx}
\usepackage{pifont} 
\usepackage{colortbl}
\usepackage{float}
\usepackage{hyphenat}

\usepackage{booktabs}
\usepackage{array}
\usepackage{wrapfig}
\usepackage{fancyvrb}

\usepackage{setspace}
\definecolor{swecream}{HTML}{EFFEFF}
\definecolor{issueborder}{HTML}{15071A}
\definecolor{issuefill}{HTML}{F6F8FA}
\definecolor{envfill}{HTML}{F2F9FF}
\definecolor{envborder}{HTML}{123C7C}
\definecolor{agentfill}{HTML}{F9F3F3}
\definecolor{agentborder}{HTML}{9B0A0A}
\definecolor{goldpatchborder}{HTML}{FABB00}
\definecolor{goldpatchfill}{HTML}{FFF7E1}

\DefineVerbatimEnvironment{CodeVerbatim}{Verbatim}{
  formatcom={\color{black}},
  fontsize=\small,
  fontfamily=\ttdefault,
  fontseries=\mddefault,
  fontshape=\updefault,
  fillcolor=\color{white},
  framerule=0pt,
}
\usepackage{longtable}
\newtcolorbox{observationbox}[1][]{
        colback=envfill,
        colbacktitle=envfill,
        colframe=envborder,
        arc=5pt,
        fontupper=\small,
        fonttitle=\bfseries\color{black},
        boxrule=0.5mm,
        boxsep=1mm,
        width=\linewidth,
        breakable,
        title={Observation \hfill #1},
        rounded corners,
        toptitle=0.7mm,
        bottomtitle=0.7mm
}
\newtcolorbox{goldpatchbox}[1][]{
        colback=goldpatchfill,
        colbacktitle=goldpatchfill,
        colframe=goldpatchborder,
        arc=5pt,
        fontupper=\small,
        fonttitle=\bfseries\color{black},
        boxrule=0.5mm,
        boxsep=1mm,
        width=\linewidth,
        breakable,
        title={\twemoji{1f6a9} Flag Captured \hfill #1},
        rounded corners,
        toptitle=0.7mm,
        bottomtitle=0.7mm
}
\newtcolorbox{issuebox}[1][]{
        colback=issuefill,
        colbacktitle=issuefill,
        colframe=issueborder,
        arc=5pt,
        fontupper=\small,
        fonttitle=\bfseries\color{black},
        boxrule=0.5mm,
        boxsep=1mm,
        width=\linewidth,
        breakable,
        title={Issue \hfill #1},
        rounded corners,
        toptitle=1mm
}
\newtcolorbox{agentbox}[1][]{
        colback=agentfill,
        colbacktitle=agentfill,
        colframe=agentborder,
        arc=5pt,
        fontupper=\small,
        fonttitle=\bfseries\color{black},
        boxrule=0.5mm,
        boxsep=1mm,
        width=\linewidth,
        breakable,
        title={EnIGMA \hfill #1},
        rounded corners,
        toptitle=1mm,
        lower separated=false
}
\newtcolorbox{fileviewerbox}[1]{
        enhanced,
        breakable,
        boxrule = 1.5pt,
        fontupper = \small,
        fonttitle = \bf\color{black},
        arc = 5pt,
        rounded corners,
        colframe = black,
        colbacktitle = swecream,
        colback = swecream,
        title = #1,
        left=4pt 
}
\newtcolorbox{promptbox}[1]{
    enhanced,
    breakable,
    boxrule=1pt,  
    fontupper=\small,
    fonttitle=\bfseries\color{black},
    arc=3pt,  
    rounded corners,
    colframe=black,
    colbacktitle=swecream,
    colback=swecream,
    title=#1,
    left=2mm,  
    right=2mm,  
    top=1mm,  
    bottom=1mm  
}

\newcommand{\cmark}{\textcolor{green!60!black}{\ding{51}}}
\newcommand{\xmark}{\textcolor{red}{\ding{55}}} 
\definecolor{revisionblue}{RGB}{0,00,255}
\newcommand{\rev}[1]{#1}
\def\BibTeX{{\rm B\kern-.05em{\sc i\kern-.025em b}\kern-.08em
    T\kern-.1667em\lower.7ex\hbox{E}\kern-.125emX}}
\begin{document}
\bstctlcite{IEEEexample:BSTcontrol}


\title{MS-Exam-Gen: An Automatic Benchmark Construction Toolkit for Evaluating LLMs on Multiple Sclerosis-MRI Imaging Knowledge}


\author{
\IEEEauthorblockN{
Abdul Basit, Muhammad Abdullah Hanif, Muhammad Shafique}
\IEEEauthorblockA{\textit{eBRAIN Lab, Division of Engineering, New York University Abu Dhabi (NYUAD)}, Abu Dhabi, UAE}
\{abdul.basit, mh6117, muhammad.shafique\}@nyu.edu
\vspace{-20pt}
}

\maketitle

\begin{abstract}
Biomedical large language model (LLM) evaluation requires \rev{auditable assessment of} narrow, evolving, source-grounded subspecialty knowledge. Multiple sclerosis MRI (MS-MRI) provides a high-stakes textual-knowledge test case because correct reasoning requires current diagnostic criteria, standardized acquisition and reporting knowledge, longitudinal monitoring concepts, lesion morphology, and recognition of difficult mimics. We present \textit{MS-Exam-Gen}, a reproducible framework for constructing and auditing a text-based multiple-choice question (MCQ) benchmark for MS-MRI knowledge; it does not evaluate direct MRI image interpretation. MS-Exam-Gen targets source-grounded criteria, protocols, reporting, and differential diagnosis. The framework combines expert-source indexing, exam-oriented topic induction, evidence-grounded MCQ generation, automated quality audits, \rev{a same-family consistency screen}, and empirical calibration. From a 66-source corpus indexed into 4,289 retrieval chunks, the pipeline produced a locked 3,058-item candidate benchmark spanning 16 topics and 53 subtopics. Evaluation across \rev{12 primary LLM endpoints yielded 36,696} item-level predictions and separated performance over a 42.8-percentage-point accuracy range (89.7\% to 46.9\%). \rev{Across these endpoints, 25.5\% of items were missed by at least four. Post-generation audits showed that refreshed construction reduced measurable answer cues, while option-order testing showed that absolute MCQ scores remain position-sensitive.} Generated construction labels remain metadata rather than validated psychometric \rev{categories}. Because expert adjudication and \rev{full option-order counterbalancing} remain future work, MS-Exam-Gen is not a clinically certified examination. It should be interpreted as an automatically filtered, source-grounded candidate benchmark and reproducible audit workflow for item-level and topic-specific LLM evaluation.
\end{abstract}


\section{Introduction}
\label{sec:intro}

Large language models (LLMs) increasingly perform well on broad medical examinations and biomedical question-answering benchmarks~\cite{moor2023foundation,Singhal2023,Thirunavukarasu2023,nori2023capabilitiesgpt4medicalchallenge}. Such aggregate scores are useful, but they provide limited evidence of subspecialty auditability or workflow-specific knowledge. Broad resources such as MedQA, PubMedQA, and MMLU-Medical do not establish whether a model can be audited against the narrow, evolving, document-grounded knowledge used in a specific clinical workflow~\cite{jin2020diseasedoespatienthave,jin2019pubmedqadatasetbiomedicalresearch,hendrycks2021measuringmassivemultitasklanguage}. For biomedical and health informatics, the evaluation problem increasingly includes not only whether an LLM ``knows medicine'' in the aggregate, but whether it can be traced against explicit criteria, protocols, source evidence, and exceptions.

Multiple sclerosis MRI (MS-MRI) textual knowledge is a demanding test case. Correct reasoning requires knowledge of dissemination-in-space/time rules, lesion morphology and distribution, standardized acquisition protocols, longitudinal activity assessment, treatment-safety surveillance, and clinically important mimics. The relevant knowledge is also evolving, with diagnostic criteria and MRI recommendations periodically revising how MS-MRI findings should be interpreted and reported~\cite{Thompson2018,Wattjes2021,Montalban2025}. Our focus is textual reasoning over MS-MRI criteria, protocols, reporting, monitoring, and mimics, not direct visual interpretation of MRI scans.

Existing benchmarks leave a gap. Broad medical QA benchmarks are too general to isolate failures in MS-MRI guideline and reporting knowledge, while MS imaging benchmarks primarily evaluate lesion segmentation or image analysis~\cite{commowick:hal-03358968,CARASS201777}. Neither setting tests whether LLMs can be audited against source-grounded textual knowledge about MS-MRI criteria, acquisition protocols, reporting standards, longitudinal monitoring, and differential diagnosis. This gap matters beyond MS: subspecialty benchmark construction is a core requirement for accountable biomedical AI evaluation when the target knowledge is narrow, evolving, and source-dependent.

Constructing such a benchmark manually is expensive, but fully automatic generation is risky unless the construction process is auditable and explicitly limited to candidate items. Medical multiple-choice questions (MCQs) can contain malformed schemas, ambiguous answer keys, superficial option-length cues, stale source assumptions, answer-label artifacts, and generated construction labels that do not match empirical difficulty. LLM-as-judge methods and model-written evaluations provide useful scalability~\cite{perez2022discoveringlanguagemodelbehaviors,zheng2023llmjudge,liu2023geval}, but they cannot replace expert clinical validation. A credible benchmark-construction workflow must therefore expose source provenance, reject malformed or cueable items, preserve raw model responses, and distinguish construction-time labels from model-panel empirical calibration. \rev{A robust workflow should also quantify residual cueing and evaluation sensitivity rather than equating automated acceptance with validation.}

We present \textit{MS-Exam-Gen}, a reproducible framework for constructing and auditing a source-grounded candidate benchmark for MS-MRI textual knowledge. MS-Exam-Gen indexes an expert corpus, induces an exam-oriented topic taxonomy using seeded retrieval, generates source-grounded MCQs, applies schema, source-evidence, redundancy, and anti-cue audits with \rev{a same-family consistency screen}, and evaluates models with stable item identifiers. The resulting locked candidate benchmark contains 3,058 MCQs spanning 16 topics and 53 subtopics. We evaluate 13 LLMs to assess discriminative utility and estimate item-level empirical difficulty from model-panel disagreement.

This work makes the following novel contributions:
\begin{enumerate}
    \item A retrieval-grounded benchmark-construction pipeline for transforming MS-MRI documents into an auditable MCQ candidate set with source provenance.
    \item An auditable quality-control protocol that records evidence coverage, schema validity, redundancy, anti-cue filtering, acceptance decisions, and raw model-response traces.
    \item A \rev{heterogeneous 12-endpoint primary calibration showing broad fixed-bank separation, topic-resolved variation, and the need to evaluate generated difficulty labels with post-generation cue and option-order analyses}.
\end{enumerate}
We frame the resulting resource as an automatically filtered candidate set, not a clinically certified examination, because no human expert item review was available.

\section{Background and Related Work}
\label{sec:related-work}

\begin{table*}[b]
\centering
\caption{Positioning of MS-Exam-Gen relative to related benchmark families.}
\label{tab:related-positioning}
\scriptsize
\setlength{\tabcolsep}{2.6pt}
\renewcommand{\arraystretch}{1.10}
\begin{tabularx}{\textwidth}{p{0.115\textwidth}p{0.225\textwidth}p{0.145\textwidth}Xccc}
\toprule
\textbf{Family} & \textbf{Representative examples} & \textbf{Useful for} & \textbf{Limitation for this study} & \shortstack{\textbf{MS-MRI}\\\textbf{text}} & \shortstack{\textbf{MS-MRI}\\\textbf{evidence}} & \shortstack{\textbf{MCQ audit}\\\textbf{/ calib.}} \\
\midrule
Broad medical QA & MedQA, PubMedQA, MMLU, MedMCQA, and MultiMedQA~\cite{jin2020diseasedoespatienthave,jin2019pubmedqadatasetbiomedicalresearch,hendrycks2021measuringmassivemultitasklanguage,pal2022medmcqa,Singhal2023} & Broad text/exam QA & Not organized around MS-MRI criteria, protocols, reporting, or mimics. & \xmark & Partial & \xmark \\
Clinical/multimodal suites & MultiMedQA-style and broader medical benchmark suites~\cite{Singhal2023,yan2024medicalbenchmarksurvey} & Clinical QA and scenarios & Coverage and provenance vary; usually not a locked MS-MRI source-grounded MCQ bank. & \xmark & Variable & Variable \\
Radiology VQA/report & Radiology foundation-model and report-generation benchmarks~\cite{wu2023radfm,liu2024radiology3dreport} & Image, VQA, reports & Image/report focused; not designed for guideline-source textual MCQ auditing. & \xmark & Partial & Variable \\
MS imaging tasks & MSSEG and longitudinal lesion segmentation~\cite{commowick:hal-03358968,CARASS201777} & Lesion segmentation & Domain-specific imaging tasks, but image labels/masks rather than textual source evidence. & \xmark & Partial & \xmark \\
LLM-generated evals & Model-written, MT-Bench, G-Eval, LLM-as-judge methods; MCQ-bias analyses~\cite{perez2022discoveringlanguagemodelbehaviors,zheng2023llmjudge,liu2023geval,shi2024judgingjudges,zheng2023mcqbias} & Generation and judging & Scalable, but source grounding and clinical safeguards are variable. & \xmark & Variable & Variable \\
\textbf{MS-Exam-Gen} & This work & Text MCQ construction & Candidate benchmark only; expert adjudication and \rev{full} option-order \rev{counterbalancing} remain future work. & \cmark & \cmark & \cmark \\
\bottomrule
\end{tabularx}
\parbox{\linewidth}{\vspace{2pt} \scriptsize\emph{Note:} Entries indicate whether a feature is a primary design goal of the family, not whether it is absent from every individual dataset. \cmark{} and \xmark{} denote yes/no; Partial and Variable denote support not specific to source-grounded MS-MRI textual MCQs.}
\end{table*}

\textbf{Medical and Clinical LLM Benchmarks:}
Medical LLM evaluation commonly uses broad examination or biomedical QA datasets, including MedQA~\cite{jin2020diseasedoespatienthave}, PubMedQA~\cite{jin2019pubmedqadatasetbiomedicalresearch}, MMLU~\cite{hendrycks2021measuringmassivemultitasklanguage}, MedMCQA~\cite{pal2022medmcqa}, and MultiMedQA-style evaluations~\cite{Singhal2023}. Recent surveys show that medical benchmarks now span text, EHR, dialogue, imaging, and multimodal tasks~\cite{yan2024medicalbenchmarksurvey}. These resources are essential for measuring broad medical knowledge. Some include expert-authored, exam-derived, or context-supported items, but they are not organized around a subspecialty imaging corpus, do not provide item-level MS-MRI source evidence or provenance, and cannot isolate failures in diagnostic criteria, MRI protocol design, longitudinal monitoring, or MS mimics.

\textbf{Radiology and MS Imaging Evaluation:}
Radiology-specific work increasingly evaluates multimodal foundation models, visual question answering, and report-oriented tasks~\cite{wu2023radfm,liu2024radiology3dreport}. MS imaging challenges such as MSSEG and longitudinal lesion segmentation provide rigorous image-analysis benchmarks~\cite{commowick:hal-03358968,CARASS201777}. These resources address important but different questions. They evaluate segmentation, visual understanding, report generation, or general radiology capability, whereas our focus is source-grounded textual reasoning over MS-MRI guidelines, acquisition recommendations, reporting standards, and differential diagnosis.

\textbf{Automatic Benchmark Construction:}
LLM-assisted benchmark generation and LLM-as-judge evaluation can scale item construction~\cite{perez2022discoveringlanguagemodelbehaviors,zheng2023llmjudge,liu2023geval}, but medical use requires stronger safeguards. LLM judges can exhibit position or scoring biases~\cite{shi2024judgingjudges}, and multiple-choice evaluations can be sensitive to option ordering and answer-label artifacts~\cite{zheng2023mcqbias}. MS-Exam-Gen therefore treats generated MCQs as candidate items, records source evidence and raw responses, audits schema and cueing failures, and operationalizes empirical difficulty using model-panel disagreement \rev{as a provisional review-prioritization signal} rather than relying on the generator's nominal construction label.

\textbf{Positioning:}
Prior work evaluates general medical knowledge, radiology images or reports, MS lesion segmentation, or generic LLM-generated benchmarks. MS-Exam-Gen occupies a complementary point in the design space: source-grounded textual MS-MRI knowledge with candidate-item framing, item-level evidence, raw-response preservation, automated quality audits, anti-cue filtering, and empirical difficulty calibration.

\section{MS-Exam-Gen Framework}
\label{sec:methodology}

\begin{figure*}[ht]
    \centering
    \includegraphics[width=\linewidth]{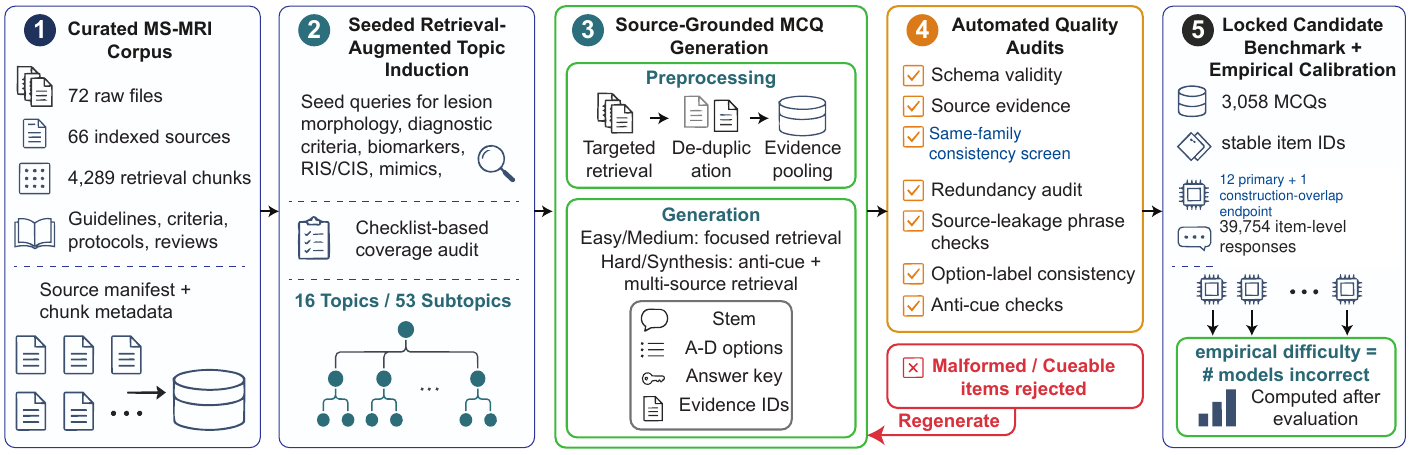}
    \caption{\textbf{MS-Exam-Gen construction and audit workflow.} Sources are indexed with metadata, sampled through seeded retrieval, organized into an exam-oriented taxonomy, used for source-grounded MCQ generation, and filtered by automated audits and \rev{a same-family consistency screen} with regeneration of malformed or cueable items. Algorithm~\ref{alg:ms-exam-gen} additionally details benchmark locking, model evaluation, and empirical calibration. Construction labels are metadata; the locked bank is a candidate benchmark, not a clinical examination.}
    \label{fig:ms_exam_gen_pipeline}
\end{figure*}

MS-Exam-Gen constructs a source-grounded MCQ benchmark candidate set from an expert corpus. \rev{Here, ``source-grounded'' means the generated content is based on retrieved evidence, with the original source and text section recorded. It does not mean that experts have confirmed the clinical claims.} The design goal is auditability across source indexing, topic induction, item acceptance/rejection, and model measurement. Figure~\ref{fig:ms_exam_gen_pipeline} and Algorithm~\ref{alg:ms-exam-gen} summarize the workflow. During the dataset construction run, taxonomy induction, item generation, JSON/schema repair, and the \rev{same-family consistency screen} used \texttt{openai/gpt-5-mini}; taxonomy and \rev{screening} calls used temperature 0.0, generation 0.3, and repair 0.1. Because GPT-5 mini is also evaluated, its score is treated as construction-overlap, not independent difficulty validation, \rev{and is excluded from primary analyses}.

\begin{algorithm}[ht]
\caption{MS-Exam-Gen construction and calibration}
\label{alg:ms-exam-gen}
\footnotesize
\begin{algorithmic}[1]
\REQUIRE Corpus $\mathcal{D}$, seed queries $\mathcal{Q}$, model panel $\mathcal{M}$
\ENSURE Locked candidate benchmark $\mathcal{B}$ with panel-dependent empirical difficulty
\STATE Build a source manifest and vector index from $\mathcal{D}$.
\STATE Retrieve passages for $\mathcal{Q}$ and induce taxonomy $\mathcal{T}$.
\FOR{each subtopic $s \in \mathcal{T}$}
    \STATE Retrieve evidence passages for $s$.
    \STATE Generate Easy/Medium items from focused evidence.
    \STATE Generate Hard and multi-source synthesis items with anti-cue constraints.
    \STATE Repair malformed JSON/schema outputs; reject unrepaired, unsupported, duplicated, or cueable items.
    \STATE \rev{Screen answerability and key consistency against stored evidence.}
\ENDFOR
\STATE Lock accepted items with stable item IDs and source provenance.
\FOR{each model $m \in \mathcal{M}$}
    \STATE Evaluate MCQs using a standardized A--D answer prompt.
    \STATE Store raw response, parsed option, parse status, and correctness by item ID.
\ENDFOR
\STATE Compute model, topic, construction-label, and empirical-difficulty summaries.
\end{algorithmic}
\end{algorithm}

\subsection{Corpus Indexing and Topic Induction}
\label{sec:topic-induction}

\rev{To make source scope and provenance auditable, }let $\mathcal{D}$ denote a curated MS-MRI corpus of consensus guidelines, diagnostic criteria, reviews, protocols, and educational sources selected for authority, recency, and concept coverage. The source manifest records inclusion status, document identifiers, metadata, and chunk counts. Documents are segmented into 1,200-character chunks with 200-character overlap; each chunk stores source path, page, start offset, ordinal, and a content-hash identifier before embedding with \texttt{text-embedding-3-large} into Chroma. MS-Exam-Gen uses \textit{seeded retrieval-augmented topic induction}: 18 seed queries target lesion location, McDonald criteria, central vein sign, paramagnetic rim lesions, RIS/CIS, NMOSD/MOGAD, pediatric/older-adult imaging, treatment surveillance, reporting, and mimics. Up to four passages per seed query, capped at 60 distinct passages, are summarized into an exam-oriented taxonomy $\mathcal{T}=\{(t_j,S_j)\}_{j=1}^{T}$, and a checklist-based coverage audit compares $\mathcal{T}$ with predefined seed concepts; this audit does not guarantee exhaustive MS-MRI topic coverage. \rev{Character-based chunking may also split tables, lists, or neighboring criteria.}

\subsection{Source-Grounded MCQ Generation}
\label{sec:mcq-generation}

\rev{To tie every candidate to retrieved evidence, }for each subtopic, the generator retrieves supporting passages and emits a four-option single-best-answer MCQ with topic, subtopic, generated construction label, question type, correct option, explanation, and source evidence. Standard Easy/Medium/Hard generation uses focused retrieval with $k=3$ chunks; the fourth stream is reported as a multi-source synthesis stratum, retrieving $k=30$ chunks and keeping at most two chunks per source and ten snippets. This stratum is stored under the generator label ``Very Hard'' in the artifact, but we avoid treating that label as validated difficulty. Options are shuffled after generation. Easy/Medium items emphasize concept recognition and interpretation, while Hard and synthesis prompts require non-recall stems, balanced option lengths, no ``most complete answer'' cues, and plausible near-miss distractors. Construction labels are metadata only and are calibrated empirically after model evaluation.

\subsection{Automated Quality Auditing}
\label{sec:validation}

\rev{To reject malformed or unsupported candidates, }each item is passed through deterministic audits before evaluation. The schema audit requires a stem, exactly four options A--D, one correct option, topic/subtopic metadata, explanation, and source evidence. Text audits flag hidden-passage phrasing, answer-label references, explanation-option contradictions after option shuffling, exact duplicates, near duplicates, and superficial trivia. Generation-time near-duplicate filters use normalized string similarity $\geq0.90$; the final audit reports remaining near-duplicate flags for transparency. Semantic duplicate checks are additionally applied to multi-source synthesis candidates, where redundancy risk is higher, using embedding cosine similarity $\geq0.90$.

\rev{To reduce superficial answer cues, }the difficulty-quality audit rejects option-length cues, excessive correct-answer completeness, weak distractors, and recall-style stems. Hard thresholds are: stem $\geq18$ words, options 8--60 words, correct-option length advantage $\leq4$ words, and correct/distractor length ratio $\leq1.15$. Multi-source synthesis thresholds are stricter: stem $\geq35$ words, options 14--70 words, advantage $\leq2$ words, ratio $\leq1.08$, and no uniquely longest correct option. These thresholds were fixed before locking the final benchmark; they are construction heuristics for removing obvious cueing artifacts, not psychometrically validated item-quality criteria. The \rev{same-family consistency screen} then receives the MCQ JSON plus stored gold context, or, when stored context is unavailable, a fresh $k=3$ retrieval context saved with the verdict, and checks answerability, key consistency, \rev{explanation consistency with stored evidence}, and uniqueness. Any major issue rejects the item; all verdicts, audit metrics, and rejection reasons are retained. Because automated judging used the same model family as generation, passing this filter is interpreted as a consistency \rev{screen conditioned on stored evidence}, not evidence of clinical correctness.

\rev{Residual cueing uses a standardized balanced logistic selector ($C=0.5$) with leave-one-topic-out folds, comparing structural option features alone and with TF--IDF stem overlap. The identical pipeline is fitted to both banks.}

\subsection{Evaluation and Empirical Difficulty}
\label{sec:benchmark}

Each locked item receives a stable identifier, and all predictions are joined by item ID rather than file position. Model outputs are parsed with conservative A--D label patterns; unparseable or exception rows are retried up to a fixed \rev{retry} limit and not cached as valid predictions. Final aggregation retains all item-model rows; parsed responses are scored, and unresolved failures are recorded as parse failures rather than silently scored as correct. For model $m$ and benchmark $\mathcal{B}$, accuracy is
\[
\mathrm{Acc}(m,\mathcal{B}) =
\frac{1}{|\mathcal{B}|}\sum_{q_i\in\mathcal{B}}
\mathbb{1}\{\hat{a}_{m,i}=a_i\},
\]
where $\hat{a}_{m,i}$ is the parsed model answer and $a_i$ is the gold option. We report overall, topic-level, generated construction-label, and empirical-difficulty results. Empirical difficulty is the number of \rev{12 non-construction-overlap endpoints} that miss an item. It is a panel-dependent calibration signal, not psychometric validation, and may reflect ambiguity, source disagreement, prompt sensitivity, or model-panel composition.

\rev{Uncertainty uses 4,000-replicate hierarchical bootstrapping over 16 construction batches and then items; a parallel analysis over 13 accuracy-independent semantic parents tests taxonomy-overlap sensitivity. Additional analyses test key disagreement and option order. The resulting review queues prioritize adjudication, and the stratified pilot does not estimate full-bank order-effect prevalence.}

\section{Experimental Setup}
\label{sec:experimental-setup}

\textbf{Corpus:}
The curated, non-exhaustive MS-MRI corpus contains 72 raw files; after excluding non-content frontmatter, the vectorstore contains 66 indexed sources and 4,289 chunks. Sources cover the 2017 McDonald criteria~\cite{Thompson2018}, 2021 MAGNIMS-CMSC-NAIMS MRI recommendations~\cite{Wattjes2021}, 2024 McDonald revisions~\cite{Montalban2025}, MS mimics, CVS/PRL, RIS/CIS, NMOSD/MOGAD, spinal/optic nerve imaging, pediatric MS, treatment surveillance, and reporting. Copyrighted source PDFs are not redistributed.

\textbf{Benchmark Construction:}
The induced taxonomy contains 16 topics and 53 exam-relevant subtopics. Standard generation produced 1,950 Easy/Medium/Hard candidates; \rev{the same-family consistency screen} accepted 1,935 (646 Easy, 646 Medium, 643 Hard). An anti-cue audit of a separate legacy Hard/multi-source synthesis pool flagged 1,217/1,293 cueable or weak items, so those items were excluded and replaced by refreshed streams. The refreshed Hard and multi-source synthesis streams each generated 900 candidates and accepted 883. The final locked benchmark contains 3,058 schema-valid MCQs: 646 Easy, 646 Medium, 883 Hard, and 883 multi-source synthesis items. For the construction-hardening audit, the 2,585-item legacy pre-refresh bank was also evaluated on the same 13-model panel. \rev{Because the legacy and locked banks are disjoint, this is a process comparison rather than an item-matched causal estimate.} The final audit found 0 schema errors, 0 exact duplicates, 54 post-hoc near-duplicate transparency flags retained because they did not meet exclusion criteria, 0 explanation-label contradictions, 0 explanation option-label references, and 0 difficulty-quality failures.

\begin{table}[t]
\centering
\caption{Construction and audit trail for the locked BHI candidate benchmark.}
\label{tab:construction-trail}
\scriptsize
\setlength{\tabcolsep}{3.0pt}
\begin{tabular}{@{}lrrr@{}}
\toprule
\textbf{Stage} & \textbf{Input} & \textbf{Accepted/flagged} & \textbf{Locked} \\
\midrule
Standard \rev{Easy/Medium} & \rev{650 each} & \rev{646 each} & \rev{646 each} \\
Standard Hard & 650 & 643 & 0 \\
Refreshed \rev{Hard/synthesis} & \rev{900 each} & \rev{883 each} & \rev{883 each} \\
Legacy Hard/synthesis pool & 1,293 & 1,217 flagged & 0 \\
\bottomrule
\end{tabular}
\parbox{\linewidth}{\vspace{1pt}\scriptsize\rev{Standard Hard and the flagged legacy pool were excluded and replaced by the refreshed strata.}}
\end{table}

\begin{table}[t]
\centering
\caption{Final benchmark and evaluation summary.}
\label{tab:benchmark-summary}
\scriptsize
\begin{tabularx}{\linewidth}{lX}
\toprule
\textbf{Component} & \textbf{Value} \\
\midrule
\rev{Corpus/taxonomy} & 66 sources; 4,289 chunks; 16 topics; 53 subtopics \\
\rev{Locked bank} & 3,058 MCQs: Easy/Medium 646 each; Hard/synthesis 883 each \\
\rev{Final audit} & 0 schema errors; 0 exact duplicates; 54 retained near-duplicate transparency flags \\
\rev{Evaluation} & 12 primary endpoints plus GPT-5 mini sensitivity; 36,696 primary/39,754 total responses; 0 parse failures \\
\bottomrule
\end{tabularx}
\end{table}

\textbf{Model Panel:}
We evaluated 13 LLMs: Gemini 2.5 Flash, GPT-5 mini, Llama 4 Maverick, DeepSeek Chat, Claude Haiku 4.5, GPT-4o mini, Gemini 2.5 Flash Lite, GPT-5 nano, Qwen3 14B, Qwen3 8B, Llama 3.1 8B, Mistral, and Phi-4 mini. GPT-5 mini is marked as construction-overlap and is not interpreted as independent benchmark validation; \rev{the other 12 endpoints form the primary panel (36,696 responses)}. Eight models were accessed through an OpenRouter-compatible interface and five through Ollama; this interface detail is reported for reproducibility and is not used to group results. All models used a standardized A--D-only prompt; temperature was set to zero where supported.

\textbf{Reproducibility Controls:}
The evaluation script stores item ID, source index, topic, subtopic, construction label, raw response, parsed answer, parse status, and correctness for every model-item pair. Interrupted runs resume from parsed checkpoints, and aggregation joins predictions by stable item ID. No unresolved unknown, unparseable, or exception rows remained in the final results. \rev{Run records preserve provider identifiers, runtime tags, decoding settings, checkpoint metadata, and model allowlists; unavailable release/quantization fields are marked ``not recorded.'' A deterministic 100-item pilot stratified by semantic parent, construction stratum, and historical disagreement used four cyclic option orders with 11 available endpoints (4,400 parsed responses); full-bank counterbalancing remains future validation.} Candidate, acceptance, rejection, and locked-retention counts are preserved.

\section{Results}
\label{sec:results}

\begin{figure}[ht]
    \centering
    \includegraphics[width=0.965\linewidth]{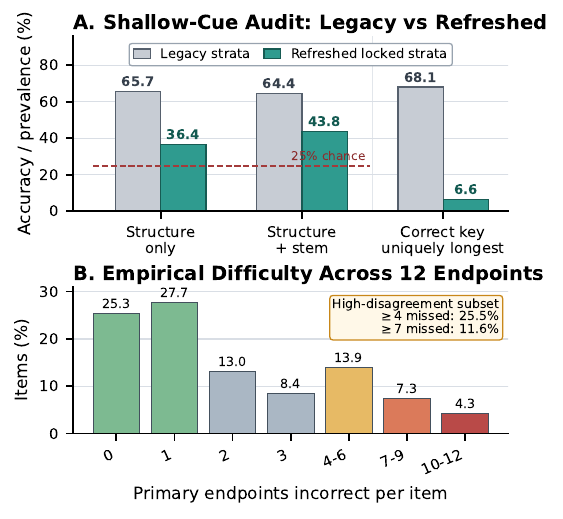}
    \caption{\rev{\textbf{Benchmark auditing and empirical calibration.} (A) The selector was fitted to the replaced legacy and locked Hard/synthesis strata. The dashed chance line applies only to the first two accuracy metrics; the third is item prevalence. Lower key predictability supports reduced measurable cueing. (B) Primary empirical difficulty is the number of 12 non-construction-overlap endpoints incorrect per item.}}
    \label{fig:calibration}
\end{figure}

\begin{figure*}[ht]
    \centering
    \includegraphics[width=0.965\linewidth]{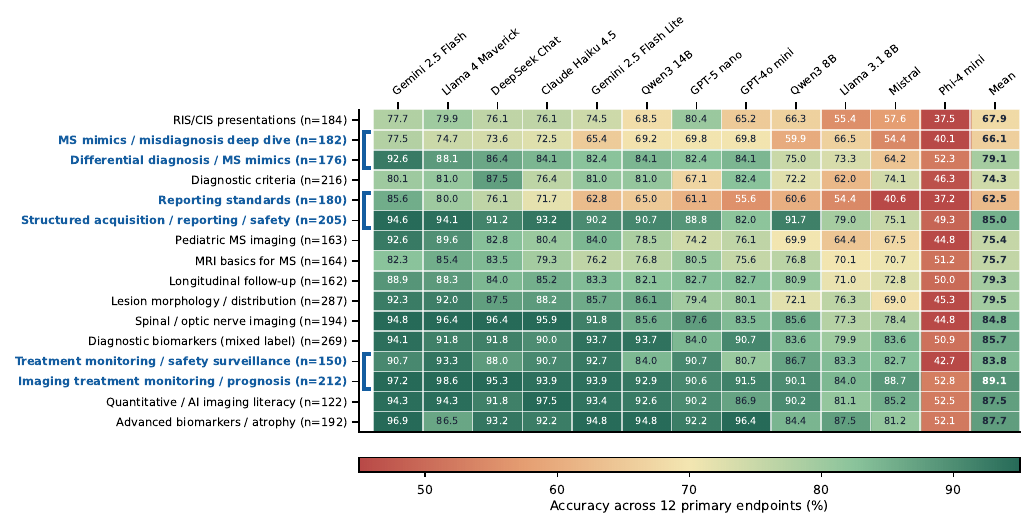}
    \caption{\rev{\textbf{Topic-level accuracy across 12 primary endpoints.} All 16 submitted topic labels and item counts are retained. Blue brackets mark three semantically overlapping label pairs and report their accuracy-independent parent means. Cells report percentages, and the final column reports each original topic's mean.}}
    \label{fig:topic-heatmap}
\end{figure*}
\vspace{-4pt}
\textbf{\rev{Overall Endpoint Separation}:}
Table~\ref{tab:model-results} shows broad \rev{performance} separation across the \rev{12 primary endpoints}, from Gemini 2.5 Flash at 89.70\% (2,743/3,058) to Phi-4 mini at 46.89\% (1,434/3,058). The intermediate range is informative: Qwen3 14B (83.32\%) approached GPT-5 nano (81.20\%), while Qwen3 8B, Llama 3.1 8B, and Mistral occupied lower but still discriminative performance bands.

\begin{table}[t]
\centering
\caption{\rev{Primary} construction-label-resolved model-panel results on the locked 3,058-item benchmark.}
\label{tab:model-results}
\scriptsize
\setlength{\tabcolsep}{1.5pt}
\renewcommand{\arraystretch}{1.03}
\resizebox{\columnwidth}{!}{%
\begin{tabular}{@{}lrrrrrl@{}}
\toprule
\textbf{Model} & \textbf{Easy} & \textbf{Med.} & \textbf{Hard} & \textbf{Synth.} & \textbf{Avg.} & \textbf{95\% CI} \\
\midrule
Gemini 2.5 Flash & \cellcolor{green!12}91.0 & \cellcolor{green!12}88.4 & \cellcolor{green!12}88.1 & \cellcolor{green!12}91.3 & \cellcolor{green!12}89.7 & \rev{86.0--92.9} \\
Llama 4 Maverick & 88.4 & 87.2 & 87.8 & 90.4 & 88.5 & \rev{85.0--91.7} \\
DeepSeek Chat & 87.8 & 86.8 & 86.2 & 87.0 & 86.9 & \rev{83.1--90.2} \\
Claude Haiku 4.5 & 86.7 & 85.1 & 81.1 & 89.5 & 85.6 & \rev{81.3--89.4} \\
Gemini 2.5 Flash Lite & 85.1 & 81.9 & 84.9 & 85.2 & 84.4 & \rev{79.2--88.9} \\
Qwen3 14B & 83.3 & 82.5 & 83.5 & 83.8 & 83.3 & \rev{78.3--87.8} \\
GPT-5 nano & 83.7 & 84.5 & 76.8 & 81.3 & 81.2 & \rev{76.5--85.5} \\
GPT-4o mini & 84.1 & 82.5 & 80.5 & 76.8 & 80.6 & \rev{75.2--85.4} \\
Qwen3 8B & 80.3 & 75.4 & 75.8 & 79.6 & 77.8 & \rev{72.5--82.5} \\
Llama 3.1 8B & 78.5 & 73.7 & 71.3 & 70.4 & 73.1 & \rev{68.0--77.8} \\
Mistral & 73.5 & 69.7 & 74.2 & 69.8 & 71.8 & \rev{65.2--77.4} \\
Phi-4 mini & 53.3 & 47.8 & 41.9 & 46.5 & 46.9 & \rev{43.8--49.9} \\
\bottomrule
\end{tabular}}
\parbox{\columnwidth}{\vspace{2pt}\scriptsize\emph{Note:} Values are percentages. ``Synth.'' denotes the multi-source synthesis stratum. \rev{CIs use 4,000 hierarchical bootstrap replicates over 16 construction batches and then items; the parallel 13-parent analysis gave similar mean interval width (8.26 vs. 8.69 points; maximum width difference 1.24). Green cells mark column maxima.}}
\end{table}

\textbf{Generated Versus Empirical Difficulty:}
Generated construction labels are metadata, not validated psychometric strata. \rev{Figure~\ref{fig:calibration}A directly tests shallow cues in the disjoint legacy and locked Hard/synthesis strata: structure-only key prediction fell from 65.7\% to 36.4\%, structure-plus-stem prediction from 64.4\% to 43.8\%, and uniquely-longest-key prevalence from 68.1\% to 6.6\%. This supports reduced measurable cueing under the refreshed process} Hard, not synthesis, remained empirically lowest-accuracy. Excluding GPT-5 mini yielded the refreshed pattern of 81.31\%, 78.79\%, 77.67\%, and 79.29\% for Easy, Medium, Hard, and synthesis.

\textbf{Empirical Difficulty Distribution:}
Figure~\ref{fig:calibration}B reports the item-level disagreement distribution. Most items \rev{showed low panel disagreement}: \rev{1,622/3,058 (53.0\%)} were missed by zero or one \rev{primary endpoint}. However, \rev{781 items (25.5\%)} were missed by at least four and \rev{355 (11.6\%)} by at least seven, forming a high-disagreement subset for error analysis and prompt-sensitivity studies. \rev{Leaving out any endpoint retained miss-count correlation $\geq0.934$ and $\geq96.2\%$ agreement on the four-endpoint flag.}

\textbf{\rev{Topic-Level Variation}:}
The \rev{lowest-accuracy item groups} were not simply the most technically advanced imaging topics. \rev{Figure~\ref{fig:topic-heatmap} displays the complete result grid. Accuracy-independent parent means were lowest for RIS/CIS (67.9\%), mimics/differential diagnosis (72.5\%), diagnostic criteria (74.3\%), and reporting/acquisition/safety (74.5\%).}

\textbf{Quality and Audit Outcomes:}
The locked benchmark is schema-valid and auditable, and the compiled evaluation contains \rev{39,754 total model-item responses (36,696 from the 12 primary endpoints)} with zero unresolved parse failures. The key negative audit result is methodological: a legacy Hard/multi-source synthesis pool contained many cueable items, and 1,217 were flagged before refreshed generation. This audit trail strengthens the locked candidate benchmark by showing that \rev{multiple construction-quality risks were tested rather than assumed absent}. \rev{The locked full bank nevertheless remained above chance for structure-only (38.3\%) and structure-plus-stem (41.2\%) key prediction, so it is not cue-free.}

\textbf{Key-Risk and Option-Order Sensitivities:}
\rev{The 7/12-majority non-key queue contains 202 items. A stricter queue contains 55 items for which five prespecified, non-overlap model families (Gemini 2.5 Flash, Llama 4 Maverick, DeepSeek Chat, Claude Haiku 4.5, and GPT-4o mini) selected the same non-key option. Removing the queues changes scores by 1.67--5.02 and 0.36--1.64 points, respectively, without erasing broad separation. The option-order pilot retains coarse ordering ($\rho=0.991$), but mean absolute score change is 2.82 points (maximum 9.0), semantic stability is 56.9\%, and 9/11 endpoints show position effects.}

\section{Discussion}
\label{sec:discussion}

\textbf{What the Benchmark Measures:}
The results should be interpreted as \rev{performance on source-grounded MS-MRI textual MCQs}, not direct image interpretation or clinical decision authority. This scope is deliberate: it isolates guideline knowledge, criteria application, reporting language, mimic recognition, and monitoring concepts. Strong performance therefore does not imply radiologist-level visual reasoning; \rev{lower performance indicates difficulty on source-derived MCQs covering MS-MRI criteria, surveillance recommendations, or differential diagnosis}. \rev{Without human or free-response baselines, scores are not measures of clinical competence or human-relative difficulty.}

\textbf{Why Construction Labels Need Calibration:}
The main methodological lesson is that generation-time difficulty is not sufficient. In the legacy bank, multi-source synthesis items were the easiest generated stratum; after refreshed anti-cue generation, Hard and synthesis accuracies fell by 4.6 and 6.4 percentage points, respectively. \rev{The disjoint-bank score changes are descriptive The identical shallow-cue diagnostic applied to both banks more directly supports reduced measurable cueing.} This supports the hardening audit but not a psychometric ordering claim: the refreshed Hard stream remained the lowest-accuracy subset, whereas the multi-source synthesis stratum was not consistently harder. We therefore report generated construction labels as metadata and use the number of models incorrect per item as the empirical calibration signal.

\textbf{Why Retain the Synthesis Stratum:}
The synthesis stratum is retained because it captures multi-source construction rather than validated difficulty. These items are source-grounded, schema-valid, and cover synthesis topics, while their higher-than-Hard accuracy demonstrates why the released artifact should pair construction labels with empirical disagreement annotations. \rev{Source count differs from Hard (7.96 vs. 2.24 per item) but is weakly associated with accuracy ($\rho=0.054$).}

\textbf{Value of a Heterogeneous Model Panel:}
Including models with varied \rev{performance profiles} reduced ceiling effects: adding GPT-5 nano, Gemini 2.5 Flash Lite, Qwen3 14B/8B, Llama 3.1 8B, Mistral, and Phi-4 mini widened the observed range to 46.9--89.7\% and exposed more discriminative items. This calibration panel helps identify \rev{performance bands and lower-accuracy content groups} rather than only confirming that strong models answer canonical guideline questions. \rev{Seven API-served endpoints retain a 9.09-point range.}

\textbf{\rev{Interpretation of Lower-Accuracy Item Groups}:}
The \rev{lower-accuracy item groups cover substantively important content areas}. Reporting standards require structured language; MS mimics require exclusionary reasoning; RIS/CIS presentations require handling incomplete clinical evolution; and diagnostic criteria require applying dissemination rules, biomarkers, and exceptions without overgeneralizing. \rev{All 16 labels are retained, while three overlapping pairs and their parent means are marked. These descriptive differences prioritize review rather than establish intrinsic clinical difficulty.}

\textbf{Audit Lessons for Automatic Benchmark Construction:}
The construction process illustrates why schema validation alone is insufficient. The initial Hard and synthesis pools contained formally valid but cueable items, and the complete legacy rerun showed inflated accuracy on the synthesis stratum. Refreshed prompts and anti-cue filters improved this problem, yet empirical results still showed that stricter prompts do not guarantee psychometric ordering. \rev{Residual cues, key-risk queues, and order effects show that automated screens do not certify validity.} A robust pipeline must preserve rejected items, report acceptance and rejection rates, and evaluate the locked bank on a heterogeneous model panel before making difficulty claims.

\textbf{Threats to Validity:}
The benchmark is automatically filtered and source-grounded but not adjudicated by human MS imaging experts, so answer-key errors, ambiguous distractors, or clinically weak items may remain. The present evidence supports use as an auditable candidate item bank, not as a clinically validated examination; expert adjudication is required before educational or clinical deployment. We also did not compare LLM performance with MS trainees, radiologists, neurologists, or other human readers; model accuracy should therefore not be interpreted relative to human expertise. The corpus is broad but not exhaustive. The evaluated models were accessed through different provider/runtime stacks, so absolute comparisons should be read as benchmark observations rather than controlled architectural claims. Finally, although options were shuffled after generation and label/explanation artifacts were audited, locked evaluations were not systematically counterbalanced across option order.

\textbf{Prospective Validation Roadmap:}
The next version should add expert adjudication across topics and empirical-difficulty bins, option-order counterbalancing, and a domain-transfer pilot in another neuroimaging subspecialty. Future releases should separate a stable core set, a high-disagreement stress-test set, and refreshed items tracking evolving criteria and biomarkers.

\section{Conclusion}
\label{sec:conclusion}
We introduced \textit{MS-Exam-Gen}, a reproducible framework for constructing and auditing source-grounded subspecialty MCQ benchmarks, demonstrated on MS-MRI knowledge. The final candidate benchmark contains 3,058 items across 16 topics and 53 subtopics, and evaluation on 13 models \rev{with 12 non-construction-overlap primary endpoints} shows a broad accuracy range from 46.9\% to 89.7\%. The results support three conclusions: \rev{an auditable subspecialty construction workflow can separate endpoint performance on a fixed bank}; topic-level analysis \rev{identifies lower-accuracy item groups for targeted review}; and generated construction labels must be calibrated empirically rather than accepted at face value. The current artifact is intended for model auditing and benchmark-construction research, not for clinical education or credentialing without expert adjudication. Future work will add expert adjudication, \rev{full} option-order counterbalancing, domain-transfer experiments, and stronger prospective validation for clinical education and model auditing.

{\footnotesize
\bibliographystyle{IEEEtran}
\bibliography{cite}
}

\end{document}